\documentclass[a4paper,conference]{IEEEtran}
\IEEEoverridecommandlockouts

\usepackage{times}  %Required
\usepackage{helvet}  %Required
\usepackage{courier}  %Required
\usepackage{url}  %Required
\usepackage{graphicx}  %Required

\usepackage{amssymb}
\usepackage{amsmath}
\usepackage{algorithm}
\usepackage[noend]{algorithmic}

\usepackage[inline]{enumitem}
\usepackage{subfig}

\usepackage{color}

\newcommand{\norm}[1]{\left\lVert#1\right\rVert}

\ifCLASSINFOpdf
\else
\fi
\begin{document}
\title{ Probabilistic Sparse Subspace Clustering Using Delayed Association \thanks{This work was supported in part by the National Science Foundation under grants IIS-1212948 and DMS-1712977   }}

\author{\IEEEauthorblockN{Maryam Jaberi}
\IEEEauthorblockA{Department of Computer Science\\
University of Central Florida\\
Orlando, FL, USA
%Email: http://www.michaelshell.org/contact.html
}
\and
\IEEEauthorblockN{Marianna Pensky}
\IEEEauthorblockA{Department of Mathematics\\
University of Central Florida\\
Orlando, FL, USA}
\and
\IEEEauthorblockN{Hassan Foroosh}
\IEEEauthorblockA{Department of Computer Science\\
University of Central Florida\\
Orlando, FL, USA}
}

%Telephone: (800) 555--1212, Fax: (888) 555--1212}


\maketitle

%%%%%%%%%% ABSTRACT

%
%%%%%%%%%% ABSTRACT
\begin{abstract}
Discovering and clustering subspaces in high-dimensional data is a fundamental problem of machine learning with a wide range of applications in data mining, computer vision, and pattern recognition. Earlier methods divided the problem into two separate stages of finding the similarity matrix and finding clusters. Similar to some recent works, we integrate these two steps using a joint optimization approach.  We make the following contributions:
(i) we estimate the reliability of the cluster assignment for each point before assigning a point to a subspace. We group the data points into two groups of ``certain'' and ``uncertain'', with the assignment of latter group delayed until their subspace association certainty improves. 
(ii) We demonstrate that delayed association is better suited for clustering subspaces that have ambiguities, i.e. when subspaces intersect or data are contaminated with outliers/noise. (iii) We demonstrate experimentally that such delayed probabilistic association leads to a more accurate self-representation and final clusters. The proposed method has higher accuracy both for points that exclusively lie in one subspace, and those that are on the intersection of subspaces. (iv) We show that delayed association leads to huge reduction of computational cost, since it allows for incremental spectral clustering. 
\end{abstract}
%
%%%%%%%%% BODY TEXT
\section{Introduction} 
The problem of clustering high dimensional data when it is formed from a union of multiple subspaces is studied in research areas such as machine learning, computer vision, and pattern recognition. In a large variety of applications, data naturally form clusters of low dimensional subspaces. In video processing, for instance, motion trajectories are usually represented by high-dimensional vectors. Yet, they can span low-dimensional linear manifolds \cite{yan2006general}.
Also, in face/image classification, under some conditions, images lie on low-dimensional linear subspaces \cite{lee2001nine}.
Subspace clustering algorithms are designed to discover clusters in a mixture of high-dimensional vectors drawn from multiple probability distributions. The idea is that, when a subset of high dimensional data belongs to a cluster, then the points in the cluster lie in a low dimensional subspace. Several methods are proposed in this area based on algebraic \cite{huang2004minimum}, iterative \cite{tseng2000nearest}, statistical \cite{ma2007segmentation} and spectral clustering \cite{liu2010robust,liu2011latent}.  Spectral clustering methods form a similarity matrix that describes the similarity between data points, in order to cluster them. In these methods, points in subspaces are self-representative. In other words, when subspaces are independent and noiseless, by having sufficient number of points in each subspace, any point in a subspace can be represented as a linear combination of other points in that subspace.
Given  a matrix $X \in \mathbb{R}^{n \times N}$, with columns  drawn from a union of $C$ independent linear subspaces of  $\mathbb{R}^{n}$, $\{ S_k\}_{k=1}^C $ with dimensions $\{d_k \ll n \}_{k=1}^C$, any data point $x_i$ can be represented as $x_i=X_{s_{\hat{k}}}z_i$, where $ x_i \in S_k$,  $X_{s_{\hat{k}}}$ are all the data points in $S_k$ except for $x_i$, and $z_i$ is a coefficient column vector. 
 Column $z_i$ can be recovered as a sparse solution of an optimization problem. The optimal solution would include non-zero coefficients corresponding to the columns of $X_{\hat{i}}$ that are in the same subspace as $x_i$.  In a general form, and in the presence of sparse norm and bounded noise or sparse outlier entries, the optimization problem can be written as follows

\vspace*{-0.3cm}                           

\begin{gather} \label{eq:SSC_All}
\min\limits_{Z} \left( \norm{Z}_\ell  + \norm{E}_{\ell^\prime} \right) 
 \quad  \text{S.T.} \quad E =  X - XZ, \quad z_{ii} = 0,
\end{gather}
where $Z$ is a coefficient matrix, $Z_i$ is its $i$-th column and $z_{ii}$ are the diagonal elements. $E$ represents a bounded noise or sparse outliers.
In the literature, the different choices of $\norm{.}_\ell$ and $\norm{.}_{\ell^\prime}$ are studied \cite{elhamifar2013sparse,liu2011latent}.
Using $Z$, a similarity matrix is defined as: $\bar{Z} = \frac{1}{2}(|Z|+|Z^T|)$.
A clustering algorithm such as normalized cuts \cite{verma2003comparison} can then be applied to the similarity matrix to find the clusters.   
The authors of \cite{guo2015robust,li2017structured} developed unified iterative frameworks for updating the low-rank matrix $Z$ using clustering results and subsequently  finding the clusters using this new version of $Z$.
The idea is that both the sparse similarity matrix and the clusters depend on each other. Thus, an alternating method can be used in the spectral clustering step to remove noise from the similarity matrix, resulting in a more accurate estimator of the similarity matrix. This leads to more accurate clusters in the spectral clustering step. This method uses an approach similar to (\ref{eq:SSC_All}) and defines an objective function as follows:

\vspace*{-0.4cm}                           

\begin{gather} \label{eq:Min_ZQ}
%\begin{aligned}
\min\limits_{Z,Q} \left( \lambda\norm{Z}_{1,Q} +  \norm{E}_{\ell^\prime} \right)  \quad \text{S.T.}  \quad z_{ii} = 0,  
%\end{aligned}
\end{gather} 
where $\norm{Z}_{1,Q}$  depends on the clustering matrix $Q \in \{0,1\}^{N\times C}$ obtained in the previous step.
An extended version of this approach is proposed in \cite{li2017structured}, where $Q$ includes continuous real values obtained by keeping the eigenvectors associated with $C$ smallest eigenvalues of the computed Laplacian matrix. This has the advantage of including continuous real values for re-weighting the representation matrix in the next iteration.  However, this method has the disadvantage of removing less noise from the similarity matrix compared to $Q \in \{0,1\}^{N\times C}$.

In this paper, we introduce a joint optimization approach that converges to an optimal solution for self-representation of clusters in high dimensional data. The novelty of our approach is that we delay the association of a point to a cluster at a given iteration, when such association is {\em uncertain}.
Remaining points are considered {\em certain}, and clustered right away. This helps improve the accuracy of updating the elements of the similarity matrix $Z$ in the next iteration. 
At each iteration, {\em certain} points, leading to get a better representation of subspaces. Also, it allows {\em uncertain} points to be drawn closer to the correct subspace before the final assignments is made. 
Two main advantages are: (i)
We effectively combine the advantages of both hard and soft clustering, leading to more accurate representation of the points in subspaces, and hence more accurate final results, since certain points are hard-clustered, whereas for uncertain points continuous values are used for re-weighting the representation in the next iteration. (ii) This process lends itself to the possibility of using an incremental spectral clustering, which in turn leads to a huge reduction in complexity and computational time. 

% %-------------------------------------------------------------------------
\section{Proposed Method} \label{sec:ProposedM}
In the proposed joint optimization methd, we alternate between finding the coefficient matrix $Z$ and the final clustering assignments. 
Assuming the initial clustering if points are given, 
we define assignments of points to $C$ clusters by a soft clustering matrix $\Phi \in [0,1]^{N \times C}$
where   elements $\phi_{ij}$ represent the probability of point $i$ belonging to subspace $j$, so that 
$
\sum_{j=1}^C \phi_{ij} =1,\quad i=1, \ldots, N.
$
In particular, $\phi_{ij} = 1$ when point $i$ is confidently assigned to cluster $j$,  and $\phi_{ij}= 0$ when point $i$ is confidently excluded from cluster $j$. Thus, we can define the {\em association matrix} $\mathit{A}$ as $\mathit{A}  = \Phi \Phi^T$.
Elements $a_{ij}$ of matrix $\mathit{A} \in [0,1]^{N \times N}$ indicate the strength of the connection between points $i$ and $j$ in the dataset. 
If $\Phi$ were a clustering matrix with entries of zero or one only, then one would have $a_{ij}=1$ if points 
$i$ and $j$ lie in the same class and $a_{ij}=0$ otherwise. 

The sparse similarity matrix $\bar{Z}$ indicates the connection between each point and all other points in the dataset. 
On the other hand, the association matrix $\mathit{A}$ represents the relationship between clustered points. 
Hence, these two matrices are related to each other and present similar information about points in the dataset.
The association matrix $\mathit{A}$ can be used to denoise and sparsify the coefficient matrix $Z$, while taking into account the coefficient matrix $Z$ can lead to more accurate recovery of the association matrix $\mathit{A}$.
The following equation formulates the connection between $Z$ and the newly introduced matrix $\mathit{A}$:
\vspace*{-0.2cm}          
\begin{gather} \label{eq:myUpdateA_Z}
\min\limits_{Z,E,A} \left( \lambda_0\norm{Z}_{1} + \frac{1}{2}\norm{E}_{F}^{2} + \lambda_1 \norm{(\mathbf{1}-\mathit{A}) * Z}_{F}^{2} \right) \\
\text{subject to} \quad E = X - XZ,  \quad z_{ii} = 0,  \quad \text{rank}(\Phi) = C.\nonumber
\end{gather}
where $ A = \Phi \Phi^T$, $\mathbf{1}^{N \times N}$ is the matrix with all unit elements, and  $*$ is the Hadamard product. The first and second term in \eqref{eq:myUpdateA_Z}, similar to equation (\ref{eq:SSC_All}), enforce the sparsity and small errors between points and their linear representations, respectively.
The last term $\norm._F^2$ in this equation imposes connectivity between points in the same subspace and removes connectivity of points in different subspaces. Indeed, since the entries   $a_{ij}$ of  matrix $\mathit{A}$ represent the probabilities of points $i$ and $j$ being in the same cluster, the elements $1 - a_{ij}$ of matrix 
$\mathbf{1}-\mathit{A}$ are smaller and do not force the respective entries $z_{ij}$ and $z_{ji}$ of matrix $Z$ to be small.

The proposed method jointly searches for a sparse self-representation matrix $Z$ that satisfies $E=X-XZ$ and the soft subspace segmentation matrix $\Phi$ that satisfies $\text{rank}(\Phi)=C$.
In order to identify the subspace clusters and mitigate/eliminate noise and outliers from the coefficient matrix $Z$ and the association matrix $\mathit{A}$, we alternate between finding the association matrix $\mathit{A}$ and the coefficient matrix $Z$. 
For a given matrix $\mathit{A}$,  the objective function (\ref{eq:myUpdateA_Z}) is convex in $\{Z, E\}$.
Given $\{Z, E\}$, we estimate $\mathit{A}$ using spectral clustering. A main novelty of our approach is in the second step, where given the matrix $\{Z, E\}$, we generate the association matrix $\mathit{A} \in [0,1]^{N \times N}$.

% \vspace*{-0.3cm}          
%
%\subsection{Delayed Association Representation }
% \vspace*{-0.1cm}          
\textbf{Updating $ \Phi $ and $  \mathit{A}$}:
Given the similarity matrix $\bar{Z}$ and the error matrix $E$, our objective is to find the soft segmentation matrix $ \Phi $ and the association matrix $\mathit{A}$. % using equation (\ref{eq:A_Phi}).
The solution for determining $\Phi $ as the probability of assigning points to $C$ clusters can be defined as a pairwise data clustering problem
\cite{verma2003comparison,shi2000normalized}. One can find hard clusters by applying a spectral clustering algorithm such as normalized cuts. 
Given the initial hard clusters obtained from the spectral clustering algorithm, we need to determine the likelihood of a point $x_i$ belonging to each subspace $\{S_k\}_{k=1}^C$, for the purpose of computing $\Phi$. We define $\norm{{\delta_{s_k}}(\bar{z_i}) }_1$	 as the {\em degree of association} of each point $x_i$ with the subspace $\{S_k\}_{k=1}^C$.
where $\bar{z_i}$  is the $i^{th}$ column of similarity matrix $\bar{Z}$ corresponding to point $x_i$, and ${\delta_{s_k}}(\bar{z_i})$ is found by keeping all the elements of the vector $\bar{z_i}$ that are associated with subspace $S_k$, and setting the remaining elements to zero. 
A point $x_i, i=1, \cdots, N,$ is assumed to be more likely to be associated with a subspace $S_k$ if it has a higher {\em degree of association} to the subspace $S_k$ defined as:
% The probability of assigning a point $x_i, i=1, \cdots, N,$ to a subspace $S_k$ can then be 
%
\vspace*{-0.2cm}                           
\begin{equation} \label{eq:probResidual}
\begin{aligned}
{p}_{ik} = \frac{\norm{{\delta_{s_k}}(\bar{z_i}) }_1}{\norm{(\bar{z_i}) }_1} \quad  k=1, \cdots, C
\end{aligned}
\end{equation}
\vspace*{-0.3cm}                           

We build the matrix $P \in  [0,1]^{N \times C}$ with elements $p_{ik}$ being  the probabilities of assigning   point $x_i$, $i=1, \cdots, N$, to the subspace  $S_k$, $k=1, \cdots C$,   where $\sum_k p_{ik} = 1$.  
The soft subspace segmentation matrix $\Phi$ is determined as described below.
For each $i$, we denote $k_i = \arg\max\limits_{1 \leq k \leq C}\, \{p_{ik}\}$ and divide points into  {\em certain} and  {\em uncertain} using the delayed association parameter $\Omega$.
The soft clustering matrix $\Phi$ is determined by   the computed probability matrix $P$ as described below.
For each $i$, we denote $k_i = \arg\max_{1 \leq k \leq C}\, \{p_{ik}\}$ and divide points into  {\em certain} and  
{\em uncertain} by applying a threshold to the probability matrix $P$ as follows:

\vspace*{-0.2cm}                           
\begin{gather} \label{eq:Phi_tresh}
\phi_{ij}=
\begin{cases}
 1, & \text{if } j = k_i  \quad \text{and}\quad p_{i k_i} \geq \Omega \\
 0, & \text{if } j \neq k_i \quad \text{and}\quad p_{i k_i} \geq \Omega \\
% \argmax_{k=1}^{C}(p_{ik}) \neq j \quad \&   \quad \max_{k=1}^C(p_{ik})\geq \Omega\\
 p_{ij}, & \text{if }  p_{i k_i} < \Omega
% \text{otherwise}
\end{cases}
\end{gather}
\vspace*{-0.2cm}

The delayed association parameter $\Omega$ is calculated by finding the average affinity between points of a subspace using the following equation:
\vspace*{-0.2cm}  
\begin{gather} \label{eq:Omega}
\Omega=1- \frac{\sum\limits_{i \neq j} M_{ij}}{ (C-1) \sum\limits_{i=j} M_{ij}}  \quad
\text{where}  \quad M = P^T P
\end{gather}
 \vspace*{-0.4cm}          

Note that matrix $M \in \mathbb{R}^{C \times C}$  demonstrates the affinity between points in subspaces. The main diagonal of this matrix shows the correlation between points of a subspace and off-diagonal entries showing the similarity of points in different subspaces. Thus, when the similarity matrix $\bar{Z}$ turns into a block diagonal matrix, the probabilities $p_{ik}$ will be pushed through zero or one and the defined matrix $M$ turns into the identity matrix with low affinity between points of different subspaces and strong connectivity among points of a subspace. We use matrix $M$ to find the delayed association parameter $\Omega$ in each iteration. Based on the definition of $\Omega$ , when there is a high ambiguity between clustered points, matrix $M$ turns into a matrix containing all similar entries and $\Omega \approx 0$. This allows more points to be grouped as {\em uncertain} and give them the chance to find a better representation before being assigned to a subspace. On the other hand, when there is a low affinity between points of different subspaces, $M$ turns into the identity matrix and $\Omega \approx 1$, which allows more points to be grouped as {\em certain}.

Given an assignment matrix $\Phi$, we form the association matrix $\mathit{A}\in \left[ 0,1\right]^{N\times N}$ which is  a symmetric matrix. For each point $x_i$ marked as {\em certain}, $a_{ij} \in \{0, 1 \}$  shows if point $x_j$ is assigned to the same class as $x_i$.  For {\em uncertain} points $a_{ij} \in [0,1]$ represents the probability of assigning $x_i$ and $x_j$ to the same cluster.
The rationale behind the method is that, in the sparse similarity matrix $\bar{Z}$, each column $i$ includes the coefficients associated with other points used to represent the point $x_i$. 
These coefficients indicate the connection between a point $x_i$ and all other points. When a point is marked as {\em certain} in the association matrix $\mathit{A}$, we discard coefficients from other clusters by setting $a_{ij} = 0$, even when the values are large. By setting $a_{ij} = 1$ for all $i, j \in S_k$, we also improve the connection between points within the same subspace.
For an {\em uncertain} point, however, there is an ambiguity regarding the correct cluster. Using the definition in formula (\ref{eq:Phi_tresh}), we preserve all strong connections, regardless of the cluster to which it is assigned in the spectral clustering step. We include all the strongly connected points while updating the coefficient matrix $Z$ in the next iteration. This approach helps us improve the connections between points and reduce noise in the next iteration.
This process is summarized in algorithm \ref{alg:updateA}.

% \vspace*{-0.1cm}          
%
%\subsection{Subspace Sparse Representation}
% \vspace*{-0.1cm}          

\textbf{Updating $\mathbf{Z}$ and $\mathbf{E}$: }
Given the probability matrix $\Phi$ and association matrix $\mathit{A}$, we update the coefficient matrix $Z$ and error matrix $E$ in the next step by solving the optimization equation (\ref{eq:myUpdateA_Z}) with respect to $\{Z, E\}$

 \vspace*{-0.3cm}                           
\begin{gather} \label{eq:myUpdate_Z}
\min\limits_{Z,E} \left( \lambda_0\norm{Z}_{1} + \frac{1}{2}\norm{E}_{F}^{2} + \lambda_1 \norm{(\mathbf{1}-\mathit{A}) * Z}_{F}^{2} \right) \\
\text{subject to} \quad E = X - XZ,  \quad z_{ii} = 0  \nonumber
\end{gather} 

Alternating between updating matrices $\{Z, E\}$, and matrices $\{\Phi, \mathit{A}\}$, as explained, helps us remove small values in the sparse coefficient matrix $Z$ and obtain a better pairwise representation of points with less noise/outliers, and hence a more accurate clustering result.

%\alglanguage{pseudocode}
%\begin{algorithm}[h]
%\small
%\caption{Finding  clustering matrix $\Phi$}
%\label{alg:updateA}
%\begin{algorithmic}[1]
\begin{algorithm}
\small
\caption{Finding clustering matrix $\Phi$}
\label{alg:updateA}
\begin{algorithmic}[1] 
\REQUIRE {Cluster assignment, similarity matrix $\bar{Z}^{(t)}$ }
\STATE Compute matrix $P^{(t)}$  using  equation (\ref{eq:probResidual})
\STATE Compute theshold $\Omega^{(t)}$  using  equation (\ref{eq:Omega})
\FOR{ $i \in \{1, .., N\}$ } 
\STATE  $k_i = \arg\max_{1 \leq k \leq C}\, \{p^{(t)}_{ik}\}$. 
\IF [Mark $i$ as ``certain'']{$P^{(t)}_{i k_i} \geq \Omega^{(t)}$}
\FOR{$j \in \{1, .., C \}$}   
 \IF {$ j = k_i$} 
\STATE $\phi_{ij}^{(t)} =1 $
 \ELSE 
 \STATE $\phi_{ij}^{(t)} =0$
\ENDIF
\ENDFOR
\ELSE  [Mark $i$ as ``uncertain'']
 \STATE  $\phi_{ij}^{(t)} =p^{(t)}_{ij} \quad \forall j \in \{1, ..,C\}$
\ENDIF
\ENDFOR
\RETURN $\Phi^{(t)}$
\end{algorithmic}
\end{algorithm}

In the initial step of the proposed method, we set $\mathit{A} = \mathbf{1}^{N \times N}$.
This is equivalent to removing the third term in the equation (\ref{eq:myUpdateA_Z}), which converts it to equation (\ref{eq:SSC_All}). We compute the coefficient matrix $Z$ and error $E$ using (\ref{eq:SSC_All}). Then, using spectral clustering \cite{shi2000normalized}, we divide the input data $X$ into $C$ clusters.
By defining association degrees (\ref{eq:probResidual}), we form the probability matrix $P$, the soft clustering matrix $\Phi$ and the association matrix $\mathit{A}$. After this initial step, we alternate between optimizing with respect to $\{\Phi,\mathit{A}\}$ and  $\{Z, E\}$.
After updating the sparse similarity matrix $\bar{Z}$, we need to find the clusters in each iteration. Previous methods (e.g. \cite{li2017structured}) use normalized cuts and recompute the solution from scratch, with a time complexity of $O(N^{3/2})$ in the best case \cite{golub2012matrix}. Our delayed association of points allows us to resort to an incremental spectral clustering \cite{ning2010incremental} at a substantially lower computational cost, since the computed eigenvectors are updated when there are changes in the similarity matrix $\bar{Z}$. The running time of this clustering approach is close to $O(N)$ when every column of the coefficient matrix $Z$ is sparse.
In our approach, when {\em certain} points do not have any coefficient that connects them to {\em uncertain} points, they do not need to be updated in spectral clustering. Incremental clustering is applied to {\em uncertain} points and all other points that are connected to {\em uncertain} in $\bar{Z}$.
As a result of updating only {\em uncertain} points in the clustering matrix, the time complexity is drastically reduced. %Similar to \cite{ning2010incremental}, we re-initialize the spectral clustering if the numbers of changes in the similarity matrix exceeds a threshold. 
In section \ref{sec:exprmntEval}, we further show that the number of {\em uncertain} points, denoted as $\kappa(\Phi^{(t)})$,  generally decreases, which implies that the cost of incremental updating itself is reducing at each iteration. The proposed method is summarized in algorithm \ref{alg:A-Z}.% We expect that the number of {\em uncertain} points, denoted as $\kappa(\Phi^{(t)})$, reduces at every step of  algorithm \ref{alg:A-Z}.% We stop the process when there is no change in the set of {\em certain} and {\em uncertain} points.

%%%%%%%%%%%%%%%%%%%%%%%%%%%%%%%%%%%%%%%%%%%%%

\begin{algorithm}[!ht]  
\small
\caption{Probabilistic sparse subspace clustering }
\label{alg:A-Z}
\begin{algorithmic}[1] 
\REQUIRE {$X \in \mathbb{R}^{n\times N}$}%: A set of high dimensional vectors  }
\STATE \textbf{Initialization:} set $\mathit{A}^{(0)} = \mathbf{1} ^{N \times N}$
\REPEAT 
\STATE \textbf{Update $Z^{(t)}$:}   
   $ \min\limits_{Z,E} \left( \lambda_0\norm{Z}_{1} + \frac{1}{2}\norm{E}_{F}^{2} + \lambda_1 \norm{(\mathbf{1}-\mathit{A}) * Z}_{F}^{2}  \right) $
   \vspace*{-0.01cm}                           
\STATE \textbf{Set}  $\bar{Z}^{(t)}=\frac{1}{2}\left[Z^{(t)}+(Z^T)^{(t)} \right]$
\STATE \textbf{Find} clusters by incremental spectral clustering\cite{ning2010incremental} or re-initialize by spectral clustering. \cite{shi2000normalized}
\STATE \textbf{Update} $\Phi^{(t)}$   using  Algorithm \ref{alg:updateA}
\STATE \textbf{Set $\mathit{A}^{(t)}$} $=(\Phi^{(t)}) (\Phi^{(t)})^T$  
\UNTIL{ $\Phi^{(t)} = \Phi^{(t-1)}$ or $\kappa\left(\Phi^{(t)}\right) \geq \kappa\left(\Phi^{(t-1)}\right)$ \text{or} $t\geq t_{\max}$ }
\end{algorithmic}
\end{algorithm}

\begin{figure}[ht]  
%\begin{multicols}{2}
\centering
\subfloat{\includegraphics[trim=1mm 8mm 27mm 7mm, clip,height=19mm]{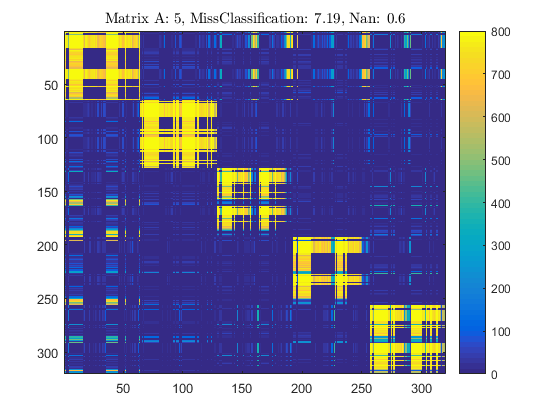}}         
%\end{subfigure}
\subfloat{\includegraphics[trim=1mm 8mm 27mm 7mm, clip,height=19mm]{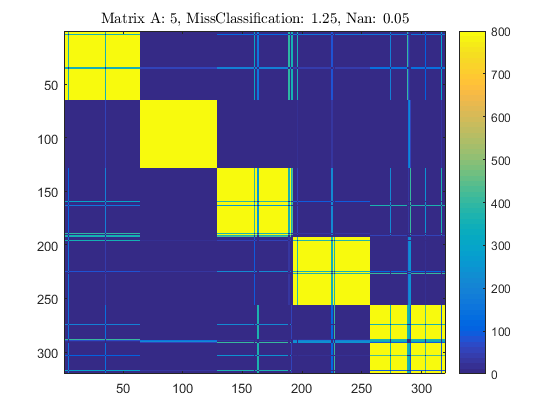}}            
\subfloat{\includegraphics[trim=1mm 8mm 1mm 7mm, clip,height=19mm]{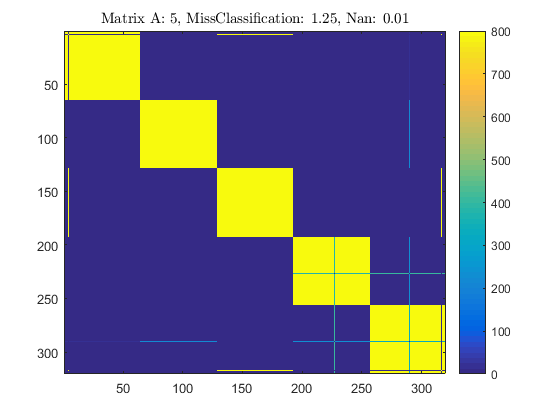}}\\
\vspace*{-0.3cm}                           
\subfloat{\includegraphics[trim = 1mm 8mm 27mm 7mm, clip,height=19mm]{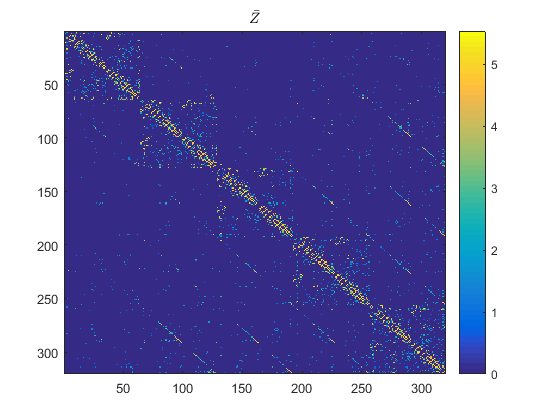}}
\subfloat{\includegraphics[trim = 1mm 8mm 27mm 7mm, clip,height=19mm]{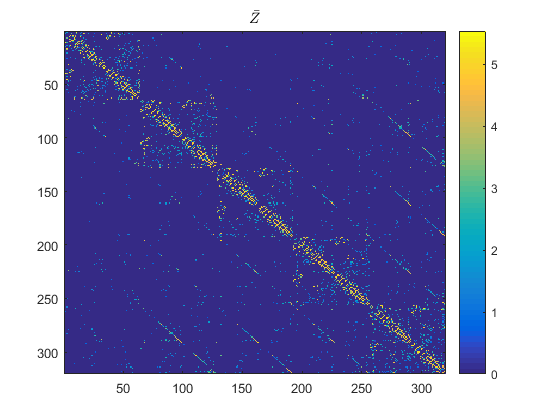}}
\subfloat{\includegraphics[trim = 1mm 8mm 1mm 7mm,  clip,height=19mm]{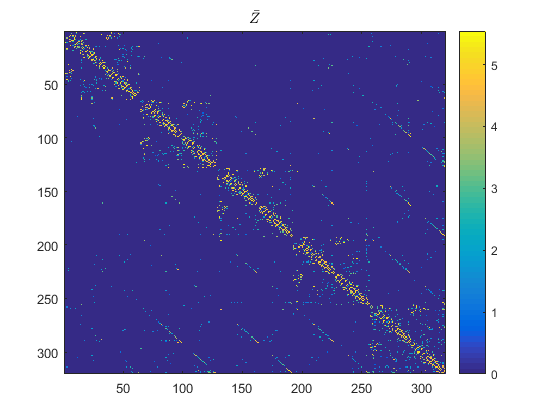}}\\
\vspace*{-0.1cm}                           
\caption{Updates in $\mathit{A}$ (top)  and  $\bar{Z}$ (bottom) over three consecutive iterations. Data is from the Yale B Face Dataset with 5 subjects. The misclassification errors are $7.19\%$, $1.25\%$ and $1.0\%$ in iterations $t=\{1, 2, 3\}$ respectively. }
 \vspace*{-0.3cm}                           

\label{fig:updateA_Z_face}
\end{figure}

As pointed out earlier, we expect that a delayed probabilistic association in subspace clustering leads us to a better self-representation matrix and better clustering assignment.
Figure \ref{fig:updateA_Z_face} illustrates an example of updates in an association matrix $\mathit{A}$ over $3$ sequential steps. The data is from the Extended Yale Database B \cite{georghiades2001few} with $C=5$ subspaces (subjects). As shown in this figure, the percentage of {\em uncertain} points is decreased from $\% \kappa_1 = 30\% $ in the first iteration to  $\% \kappa_3 = 1\% $ in the third iterations. Also, the misclassification errors of subspace clustering are decreased from $7.19\%$ to $0.1\%$. 
 \vspace*{-0.3cm}                           

%\vspace*{-0.6cm}
%-------------------------------------------------------------------------
\section{Experimental Results}\label{sec:exprmntEval}

%\subsection{Error Metrics}
 To evaluate the accuracy of the proposed subspace clustering method, we used different metrics. The first metric we used is a direct measure of the misclassification error of sparse subspace clustering results. This is defined as:

\vspace*{-0.3cm}                           

\begin{equation} \label{eq:MetricsACC}
Err = \frac{\# \mathrm{Misclassified \, Points}}{N} 
\end{equation} 
This is the total number of incorrectly clustered points over the total number of points in the population. 
%A second metric (equation (\ref{eq:MetricsACC_Nan})) measures the misclassification error of subspace clustering using only the points that are not on the intersection of subspaces, i.e. with a high probability of belonging to only one subspace.
%
%\vspace*{-0.2cm}                           
%
%\begin{equation} \label{eq:MetricsACC_Nan}
%Err_{\hat{\kappa}} = \frac{\#\mathrm{Misclassified \, Points}_{ \hat{\kappa}}}{N_{\hat{\kappa}} } 
%\end{equation} 
%where ${\#\mathrm{Misclassified \, Points}_{ \hat{\kappa}}}$ is the number of points that are not on the intersection of subspaces and are clustered incorrectly, and $N_{\hat{\kappa}}$ is the total number of points that are not on subspace intersections. 
Another metric used is the subspace sparse recovery error \cite{soltanolkotabi2012geometric}. This metric computes the error of representing each point in its final cluster according to the coefficient matrix $Z$. The columns of $z_i$ determine all the coefficients to self-represent the point $x_i$ using all the other points in the dataset. Using the result of the clustering algorithm, each column $z_i$ gets divided into $C$ classes
${\delta_{s_1}}({z_i}), {\delta_{s_2}}({z_i}), ..., {\delta_{s_C}}({z_i})$, where each of the ${\delta_{s_k}}({z_i})$ are the coefficients of representing $x_i$ in cluster $k$. Assuming $m$ is the correct cluster for point $x_i$, and that ${\delta_{s_m}}({z_i})$ are the corresponding coefficients to reconstruct the point, the average subspace sparse recovery error can be defined as:

\begin{equation} \label{eq:MetricsSSR}
SSR  = 1-\frac{1}{N} \sum_{i=1}^N  \frac{\norm{\delta_{s_m}(Z_i)}_1}{\norm{Z_i}_1} 
\end{equation} 

In the following sections, we study the accuracy of the proposed method for both synthetic and real datasets, and compare the results with state-of-the-art algorithms. In our experiments, the value of $t_{\max}=10$ and $\frac{\lambda_0}{\lambda_1} = 100$.

\subsection{Synthetic Data: Subspace Intersection}

As pointed out earlier, we expect that a delayed probabilistic association in subspace clustering is better suited to handle subspaces with large intersections and overlaps, i.e. those with a large number of points that belong to more than one subspace. This is of practical interest, since in most real data there is either ambiguity due to noise or significant similarity of points in different clusters, due to nested subspaces. In this section, we experimentally study this important problem. 
To generate our dataset, we used the method described in \cite{soltanolkotabi2012geometric}. We examine the effect of intersection between subspaces when each subspace $\{S_j\}_{j=1}^C$ has a true dimension of $d = 10$ in $\mathbb{R}^n$ with an ambient dimension $n=200$, and an intersection dimension of $s$ (i.e. sharing $s$ basis vectors). The first subspace $S_1$ of dimension $d$ is generated uniformly at random. To generate each of the remaining subspaces for each subspace $\{S_j\}_{j=2}^C$, we generated two sets of basis: (i) the intersection basis $S^{(1)}_j$ of  dimension $s$ where $s<d$. (ii) the disjoint basis $S^{(2)}_j$ with dimension $d-s$. Then, the basis for each of the remaining subspaces ($S_j$,  $j=1..C$)  are formed as $S_j = S^{(1)}_j \cup S^{(2)}_j$.
We generated three different models by varying the number of subspaces $C$ from two to four and samples $N_j =100$ points uniformly at random from each subspace. We generated $20$ instances from each of these models and changed the ratio of the intersection between subspaces in the range $\frac{s}{d} = \{40\%, .., 90\% \}$.  

\begin{figure}[ht]  
%\begin{multicols}{2}
\centering
{\includegraphics[trim = 1mm 0mm 10mm 1mm, clip,height=26mm]{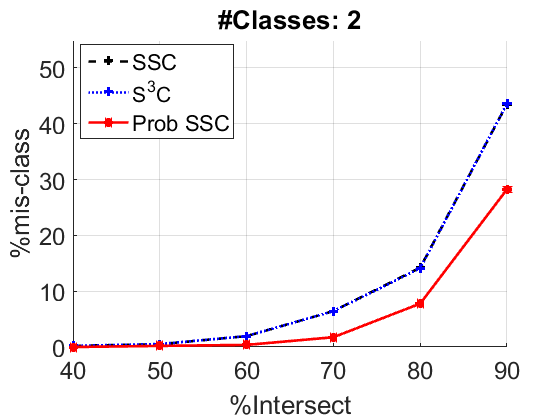}}%\hfill
%{\includegraphics[trim = 10mm 0mm 10mm 1mm, clip,height=35mm]{intersect3cls.png}}
{\includegraphics[trim = 10mm 0mm 10mm 1mm, clip,height=26mm]{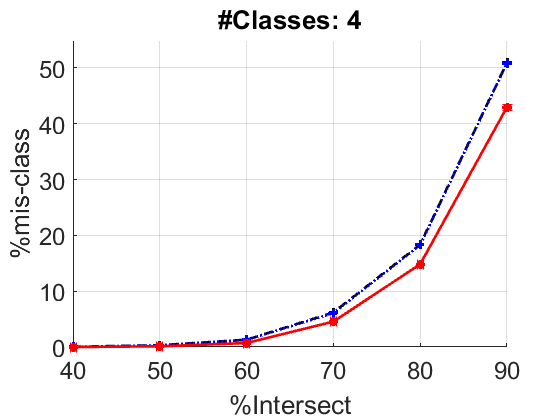}}
\vspace*{-0.2cm}                           
\caption{Average $\%$ misclassification errors (\ref{eq:MetricsACC}) for Prob-SSC,  SSC and S$^3$C methods %in \cite{elhamifar2009sparse,li2015structured} 
over $20$ independent trials.}

\label{fig:missClass_Intersect}
\end{figure}

\begin{figure}[ht]  
%\begin{multicols}{2}
\centering
{\includegraphics[trim = 10mm 0mm 10mm 1mm, clip,height=26mm]{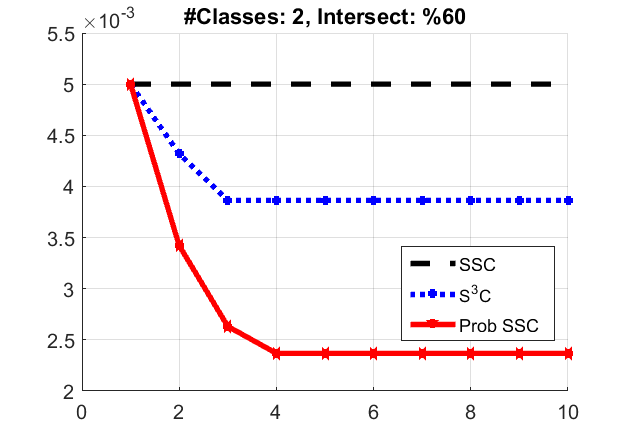}}%\hfill
%{\includegraphics[trim = 10mm 0mm 10mm 1mm, clip,height=33mm]{Intersect_3Class_prc60.png}}
{\includegraphics[trim = 10mm 0mm 10mm 1mm, clip,height=26mm]{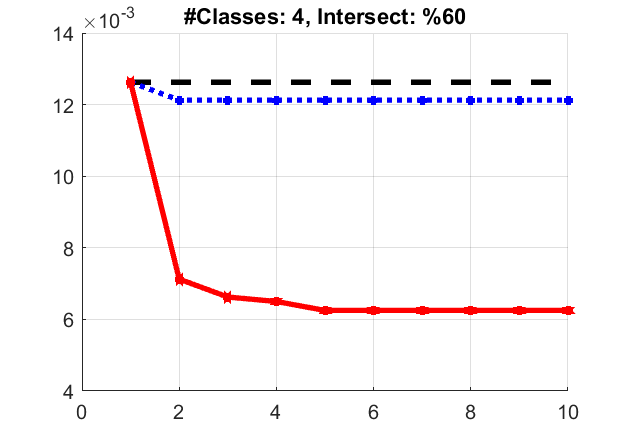}}
\vspace*{-0.3cm}                           
\caption{Average $\%$misclassification errors over 10 iterations. Results  are shown for Prob-SSC, SSC and S$^3$C.  %methods in \cite{elhamifar2009sparse,li2015structured} over $20$ independent trials.
}
\label{fig:missClass_change_Intersect}
\vspace*{-0.2cm}                           
\end{figure}

\textbf{Clustering Accuracy:} In the first experiment, we examined the convergence of the algorithm, i.e. the clustering misclassification versus the number of iterations. We show the result for $50\%$ and $90\%$ intersections and for two to four clusters. The average number of iterations $t$ before reaching the condition in algorithm \ref{alg:A-Z} is $t = 5$.
The complete summary of average misclassification error over $20$ independent trials is shown in table \ref{table:ClassIntersect_Err}. As shown in this table, we compare the misclassification error of the proposed method with SSC \cite{elhamifar2009sparse} as a baseline and S$^3$C  \cite{li2017structured} as the state-of-the-art. The proposed delayed association method consistently outperforms both the baseline and the state-of-the-art methods. 
Additionally, table \ref{table:ClassIntersect_SSR} shows a complete summary of average representation errors (equation (\ref{eq:MetricsSSR})) on the same experiment. As seen in this table, our method computes a sparser matrix $Z$ compared with SSC. However, in some cases, it may compute a less sparse matrix $Z$ compared with S$^3$C due to delayed association.
When there is more ambiguity in the clusters, our method keeps more points in the {\em uncertain} group, making the coefficient matrix $Z$ temporarily less sparse compared with S$^3$C. However, this helps keep the classification error lower (see Table \ref{table:ClassIntersect_Err}).
%%intermediate iterations
 Figure \ref{fig:missClass_Intersect} shows the average misclassification error (equation (\ref{eq:MetricsACC})) for $20$ independent trials. We compared the results of the proposed method with SSC \cite{elhamifar2009sparse} and S\textsuperscript{3}C \cite{li2017structured}.  As illustrated, the proposed method is more accurate compared with state-of-the-art algorithms.  Graphs in figure \ref{fig:missClass_change_Intersect} illustrate the average changes in accuracy of subspace clustering by alternately updating $\{Z, E\}$ and $\mathit{A}$. These graphs compare the accuracy of the proposed algorithm with the approaches in S$^3$C \cite{li2017structured} and in SSC.

\begin{table*} [ht]                                           
\centering                                             
\caption{Intersection $\%$misclassification as a function of subspaces intersection   and number of subspaces.}         
\vspace*{-0.2cm}                           
                      
\begin{tabular}{|c|c|c|c|c|c|c|c|c|c|c|c|c|c|c|c|}   
\hline                                               
\multicolumn{1}{|c}{ $\#$Subspace} &\multicolumn{5}{|c|}{$2$ } 
& \multicolumn{5}{c|}{$3$ } &\multicolumn{5}{c|}{$4$ }
\\
\multicolumn{1}{|c}{$\%$intersect} 
&\multicolumn{1}{|c}{$50\%$} 		& \multicolumn{1}{c}{$60\%$} &\multicolumn{1}{c}{$70\%$} 		& \multicolumn{1}{c}{$80\%$} &\multicolumn{1}{c|}{$90\%$} 		
&\multicolumn{1}{c}{$50\%$} 		& \multicolumn{1}{c}{$60\%$} &\multicolumn{1}{c}{$70\%$} 		& \multicolumn{1}{c}{$80\%$} &\multicolumn{1}{c|}{$90\%$} 
&\multicolumn{1}{c}{$50\%$} 		& \multicolumn{1}{c}{$60\%$} &\multicolumn{1}{c}{$70\%$} 		& \multicolumn{1}{c}{$80\%$} &\multicolumn{1}{c|}{$90\%$}  \\
\hline  
\hline  
\multicolumn{1}{|c}{SSC} 
& \multicolumn{1}{|c}{0.10} &\multicolumn{1}{c}{0.50} 		& \multicolumn{1}{c}{2.68}  &\multicolumn{1}{c}{8.0} 	&\multicolumn{1}{c|}{33.8} 	
&\multicolumn{1}{c}{0.13} 		& \multicolumn{1}{c}{1.20} &\multicolumn{1}{c}{ 3.98} 		& \multicolumn{1}{c}{13.3} &\multicolumn{1}{c|}{ 42.0} 
&\multicolumn{1}{c}{0.31} 		& \multicolumn{1}{c}{1.30} &\multicolumn{1}{c}{ 6.19} 		& \multicolumn{1}{c}{18.4} &\multicolumn{1}{c|}{50.9}\\
\multicolumn{1}{|c}{S$^3$C} 
&\multicolumn{1}{|c}{\textbf{0.03}} 		& \multicolumn{1}{c}{0.40} &\multicolumn{1}{c}{2.38} 		& \multicolumn{1}{c}{7.63} &\multicolumn{1}{c|}{30.9} 	
&\multicolumn{1}{c}{0.15} 		& \multicolumn{1}{c}{{1.10}} &\multicolumn{1}{c}{3.98} 		& \multicolumn{1}{c}{13.3} &\multicolumn{1}{c|}{ 42.0} 
&\multicolumn{1}{c}{0.30} 		& \multicolumn{1}{c}{1.23} &\multicolumn{1}{c}{ 5.94} 		& \multicolumn{1}{c}{18.1} &\multicolumn{1}{c|} {50.5} \\
\hline
\hline
\multicolumn{1}{|c}{Prob SSC} 
&\multicolumn{1}{|c}{\textbf{0.03}} 		& \multicolumn{1}{c}{\textbf{0.25}} &\multicolumn{1}{c}{\textbf{1.58}} 		& \multicolumn{1}{c}{\textbf{7.25}} &\multicolumn{1}{c|}{\textbf{24.9}} 		
&\multicolumn{1}{c}{\textbf{0.08}} 		& \multicolumn{1}{c}{\textbf{ 0.65}} &\multicolumn{1}{c}{\textbf{3.23}} 		& \multicolumn{1}{c}{\textbf{11.7}} &\multicolumn{1}{c|}{\textbf{34.8}}
&\multicolumn{1}{c}{\textbf{0.11}} 		& \multicolumn{1}{c}{\textbf{0.66}} &\multicolumn{1}{c}{\textbf{4.41}} 		& \multicolumn{1}{c}{\textbf{14.4}} &\multicolumn{1}{c|}{\textbf{42.4}}  \\
\hline
\end{tabular}                                          
\label{table:ClassIntersect_Err}                             
\end{table*}

\begin{table*} [ht]                              
\centering                                             
\caption{Intersection $\%$SSR error as a function of subspaces intersection   and number of subspaces.}    
\vspace*{-0.2cm}                           
                           
\begin{tabular}{|c|c|c|c|c|c|c|c|c|c|c|c|c|c|c|c|}   
\hline                                               
\multicolumn{1}{|c}{ $\#$Subspace} &\multicolumn{5}{|c|}{$2$ } 
& \multicolumn{5}{c|}{$3$ } &\multicolumn{5}{c|}{$4$ }
\\
\multicolumn{1}{|c}{$\%$intersect} 
&\multicolumn{1}{|c}{$50\%$} 		& \multicolumn{1}{c}{$60\%$} &\multicolumn{1}{c}{$70\%$} 		& \multicolumn{1}{c}{$80\%$} &\multicolumn{1}{c|}{$90\%$}  		
&\multicolumn{1}{c}{$50\%$} 		& \multicolumn{1}{c}{$60\%$} &\multicolumn{1}{c}{$70\%$} 		& \multicolumn{1}{c}{$80\%$} &\multicolumn{1}{c|}{$90\%$} 
&\multicolumn{1}{c}{$50\%$} 		& \multicolumn{1}{c}{$60\%$} &\multicolumn{1}{c}{$70\%$} 		& \multicolumn{1}{c}{$80\%$} &\multicolumn{1}{c|}{$90\%$}  \\
\hline  
\hline  
\multicolumn{1}{|c}{SSC} 
&\multicolumn{1}{|c}{8.03} 		& \multicolumn{1}{c}{19.3} &\multicolumn{1}{c}{22.6  } 		& \multicolumn{1}{c}{28.6} &\multicolumn{1}{c|}{34.8} 		
&\multicolumn{1}{c}{11.8 } 		& \multicolumn{1}{c}{ 17.7 } &\multicolumn{1}{c}{28.3 } 		& \multicolumn{1}{c}{ 36.0 } &\multicolumn{1}{c|}{53.4} 
&\multicolumn{1}{c}{12.5} 		& \multicolumn{1}{c}{25.9 } &\multicolumn{1}{c}{28.3} 		& \multicolumn{1}{c}{48.6} &\multicolumn{1}{c|}{60.4}  \\

\multicolumn{1}{|c}{S$^3$C}
&\multicolumn{1}{|c}{{3.64}} 		& \multicolumn{1}{c}{{ 13.3 }} &\multicolumn{1}{c}{\textbf{15.9} } 		& \multicolumn{1}{c}{\textbf{21.1}} &\multicolumn{1}{c|}{37.8} 		
&\multicolumn{1}{c}{5.17} 		& \multicolumn{1}{c}{\textbf{8.68}} &\multicolumn{1}{c}{\textbf{21.8}} 		& \multicolumn{1}{c}{\textbf{27.3}} &\multicolumn{1}{c|}{{47.9} }
&\multicolumn{1}{c}{\textbf{4.93 }} 		& \multicolumn{1}{c}{\textbf{13.2}} &\multicolumn{1}{c}{ \textbf{16.1 } 	}	& \multicolumn{1}{c}{\textbf{42.0}} &\multicolumn{1}{c|} {{57.4 }}  \\
\hline
\hline
\multicolumn{1}{|c}{Prob SSC} 
&\multicolumn{1}{|c}{\textbf{3.04}} 		& \multicolumn{1}{c}{\textbf{10.5}} &\multicolumn{1}{c}{{17.7}} 		& \multicolumn{1}{c}{{25.3}} &\multicolumn{1}{c|}{\textbf{35.7}} 		
&\multicolumn{1}{c}{\textbf{3.93}} 		& \multicolumn{1}{c}{\textbf{8.90}} &\multicolumn{1}{c}{{22.8}} 		& \multicolumn{1}{c}{{32.2}} &\multicolumn{1}{c|}{\textbf{46.3} }
&\multicolumn{1}{c}{5.81} 		& \multicolumn{1}{c}{19.2} &\multicolumn{1}{c}{19.5} 		& \multicolumn{1}{c}{ 43.6} &\multicolumn{1}{c|}{ \textbf{56.6} } \\
\hline
\end{tabular}                                          
\label{table:ClassIntersect_SSR}                             
\end{table*}  

\begin{figure}[ht]  
%\begin{multicols}{2}
\centering
{\includegraphics[trim = 1mm 0mm 10mm 1mm, clip,height=26mm]{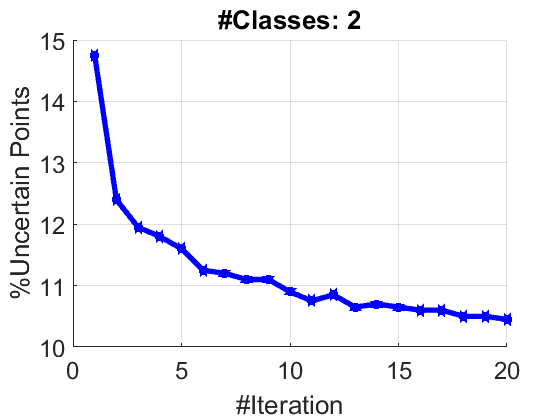}}%\hfill
{\includegraphics[trim = 10mm 0mm 10mm 1mm, clip,height=26mm]{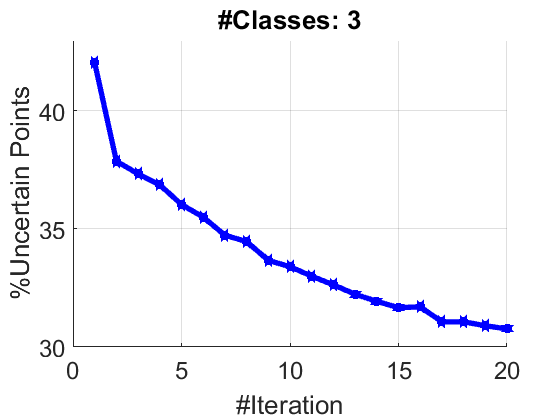}}
%{\includegraphics[trim = 10mm 0mm 10mm 1mm, clip,height=32mm]{Intersect_4Class_prc60.png}}
\vspace*{-0.3cm}                           
\caption{Average decrease of $\%${\em uncertain} points.}
\label{fig:decrease_Nan}
\vspace*{-0.2cm}                           
\end{figure}

To show that the percentage of uncertain points ($\kappa(\Phi)$) decreases and leads to converge the proposed method,  we removed  constraints of $\Phi^{(t)} = \Phi^{(t-1)}$ or $\kappa\left(\Phi^{(t)}\right) \geq \kappa\left(\Phi^{(t-1)}\right)$ in algorithm \ref{alg:A-Z} and repeated the process for $20$ iterations. Figure \ref{fig:decrease_Nan} shows the average percentage of uncertain points decreases monotonically. The experiment is on $50\%$ intersection between subspaces over 20 independent trials.

\begin{table*} [ht]                                         
\centering                                             
\caption{Average $\%$misclassification errors on Extended Yale B Dataset  \cite{georghiades2001few}.}  
\vspace*{-0.2cm}        
\begin{tabular}{|c|c|c|c|c|c|c|c|c|c|c|}   
\hline                                               
\multicolumn{1}{|c}{$\#$ Subjects} &\multicolumn{2}{|c|}{$2$ } 
& \multicolumn{2}{|c|}{$3$ } &\multicolumn{2}{|c|}{$5$ }
& \multicolumn{2}{|c|}{$8$ } &\multicolumn{2}{|c|}{$10$}
\\
\multicolumn{1}{|c}{$\%$Err} 
&\multicolumn{1}{|c}{Average} & \multicolumn{1}{c|}{Median} 
&\multicolumn{1}{|c}{Average} & \multicolumn{1}{c|}{Median} 
&\multicolumn{1}{|c}{Average} & \multicolumn{1}{c|}{Median} 
&\multicolumn{1}{|c}{Average} & \multicolumn{1}{c|}{Median} 
&\multicolumn{1}{|c}{Average} & \multicolumn{1}{c|}{Median} 
\\
\hline  \hline  
\multicolumn{1}{|c}{LRR} 
&\multicolumn{1}{|c}{6.74} 	& \multicolumn{1}{c|}{7.03} 
&\multicolumn{1}{|c}{9.30} 	& \multicolumn{1}{c|}{9.90} 
&\multicolumn{1}{|c}{13.94} & \multicolumn{1}{c|}{14.38} 
&\multicolumn{1}{|c}{25.61} & \multicolumn{1}{c|}{24.80} 
&\multicolumn{1}{|c}{29.53} & \multicolumn{1}{c|}{30.00}
\\
\multicolumn{1}{|c}{LRSC} 
&\multicolumn{1}{|c}{3.15} 	& \multicolumn{1}{c|}{2.34} 
&\multicolumn{1}{|c}{4.71} 	& \multicolumn{1}{c|}{4.17} 
&\multicolumn{1}{|c}{13.06} & \multicolumn{1}{c|}{8.44} 
&\multicolumn{1}{|c}{26.83} & \multicolumn{1}{c|}{28.71} 
&\multicolumn{1}{|c}{35.89}	& \multicolumn{1}{c|}{34.84} 
\\
\multicolumn{1}{|c}{LatLRR} 
&\multicolumn{1}{|c}{2.54} 	& \multicolumn{1}{c|}{0.78} 
&\multicolumn{1}{|c}{4.21} 	& \multicolumn{1}{c|}{2.60} 
&\multicolumn{1}{|c}{6.90} 	& \multicolumn{1}{c|}{5.63} 
&\multicolumn{1}{|c}{14.34} & \multicolumn{1}{c|}{10.06} 
&\multicolumn{1}{|c}{22.92} & \multicolumn{1}{c|}{23.59} 
\\
\multicolumn{1}{|c}{SSC} 
& \multicolumn{1}{|c}{1.87} & \multicolumn{1}{c|}{0.00} 
& \multicolumn{1}{|c}{3.35} & \multicolumn{1}{c|}{0.78} 
&\multicolumn{1}{|c}{4.32} 	& \multicolumn{1}{c|}{2.81} 
&\multicolumn{1}{|c}{5.99} 	& \multicolumn{1}{c|}{4.49} 
&\multicolumn{1}{|c}{7.29} 	& \multicolumn{1}{c|}{5.47} 
\\
\multicolumn{1}{|c}{S$^3$C }% \cite{li2015structured}} 
&\multicolumn{1}{|c}{1.27} & \multicolumn{1}{c|}{\textbf{0.00}} 
&\multicolumn{1}{|c}{2.71} & \multicolumn{1}{c|}{\textbf{0.52}} 
&\multicolumn{1}{|c}{3.41} & \multicolumn{1}{c|}{{1.25}} 
&\multicolumn{1}{|c}{4.15} & \multicolumn{1}{c|}{{2.93}} 
&\multicolumn{1}{|c}{5.16} & \multicolumn{1}{c|}{{4.22}}
\\
\multicolumn{1}{|c}{Soft S$^3$C }%\cite{li2017structured}} 
&\multicolumn{1}{|c}{0.76} & \multicolumn{1}{c|}{\textbf{0.00}} 
&\multicolumn{1}{|c}{0.82} & \multicolumn{1}{c|}{\textbf{0.52}} 
&\multicolumn{1}{|c}{1.32} & \multicolumn{1}{c|}{{1.25}} 
&\multicolumn{1}{|c}{2.14} & \multicolumn{1}{c|}{{1.95}} 
&\multicolumn{1}{|c}{2.40} & \multicolumn{1}{c|}{{2.50}}
\\                       
 \hline
 \hline
\multicolumn{1}{|c}{Prob SSC}
& \multicolumn{1}{|c}{\textbf{0.48}}	& \multicolumn{1}{c|}{\textbf{0.00}} 
& \multicolumn{1}{|c}{\textbf{0.77}}	& \multicolumn{1}{c|}{\textbf{0.52}} 
& \multicolumn{1}{|c}{\textbf{1.23}}	& \multicolumn{1}{c|}{\textbf{0.93}} 
& \multicolumn{1}{|c}{\textbf{2.08}}	& \multicolumn{1}{c|}{\textbf{1.26}}
& \multicolumn{1}{|c}{\textbf{2.14}} 	& \multicolumn{1}{c|}{\textbf{2.19}}
\\
\hline
\end{tabular}     
                                     
\label{table:face_ACC}                             
\end{table*}

\subsection{Real Data: Face Clustering }
Face classification is one of the many applications of subspace clustering. Face image classification techniques try to cluster images of the same subject under varying illumination or imaging conditions in one group.  Study in \cite{lee2001nine} shows that a set of images of an object under varying illumination lies in a low-dimensional linear subspace of the image space of up to nine dimensions. Thus, face images in this condition can be clustered using subspace clustering techniques. The Extended Yale B Database \cite{georghiades2001few} is a facial dataset widely used in subspace clustering literature \cite{elhamifar2013sparse,wright2009robust} and contains $2,414$ frontal face images of $38$ human subjects taken under approximately $64$ different illumination conditions. 
This dataset is considered as a challenging one for clustering techniques due to its extreme lighting variations. %(Figure \ref{fig:examYaleface}).
In this experiment, we used the proposed algorithm to study the improvement in accuracy of face clusters compared to previous methods. 
We used subsets of $C = \{2, 3, 5, 8, 10\}$ different subjects (subspaces) from the dataset. Each subject includes $64$ images ($N_j = 64$). Each downsampled image has a dimension of $48 \times 42$ that is vectorized to a $2016$-dimensional vector. These $38$ subjects are divided in $4$ groups of 1-10, 11-20, 21-30 and 31-38 \cite{li2017structured}. Similar to the reported results in \cite{li2017structured}, we kept the original size of the image vectors ($2016$) and reported the result on the whole dataset. 
 We examined the accuracy of the proposed subspace clustering algorithm on this dataset. 
Table \ref{table:face_ACC} shows the clustering error percentages for the proposed algorithm. We compared the error of our proposed method with S$^3$C \cite{li2017structured}, SSC\cite{elhamifar2009sparse}, and LRR \cite{liu2010robust}.%, LatLRR \cite{liu2011latent} and LRSC \cite{vidal2014low}. 
We cite the reported results in \cite{vidal2014low,liu2011latent} and  \cite{li2017structured} in this table. As it is shown in this table, our method outperforms all other state-of-the-art algorithms.

\begin{figure}[ht]
\centering
\subfloat[]{\includegraphics[height=17mm]{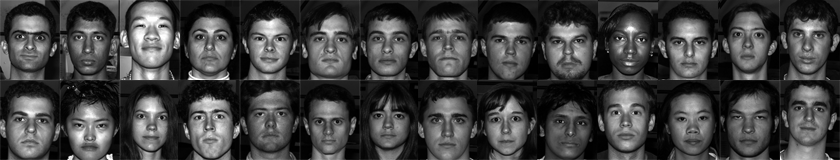}\label{fig:examYaleface}} 
\vspace*{-0.01 cm}
\subfloat[]{\includegraphics[height=13mm]{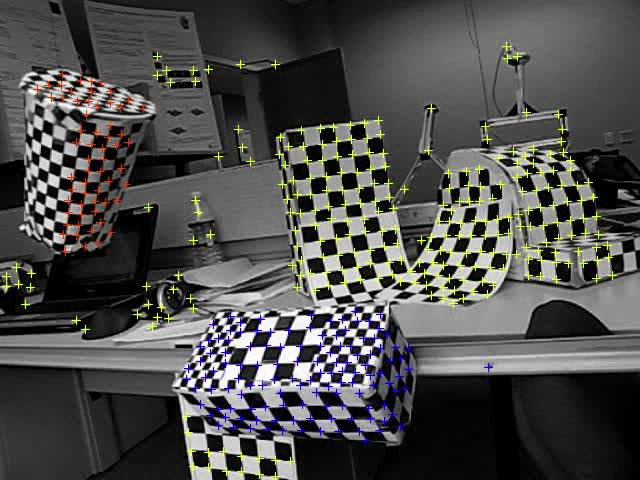} 
\includegraphics[height=13mm]{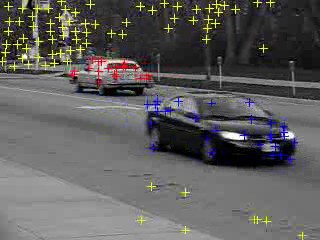}   
\includegraphics[height=13mm]{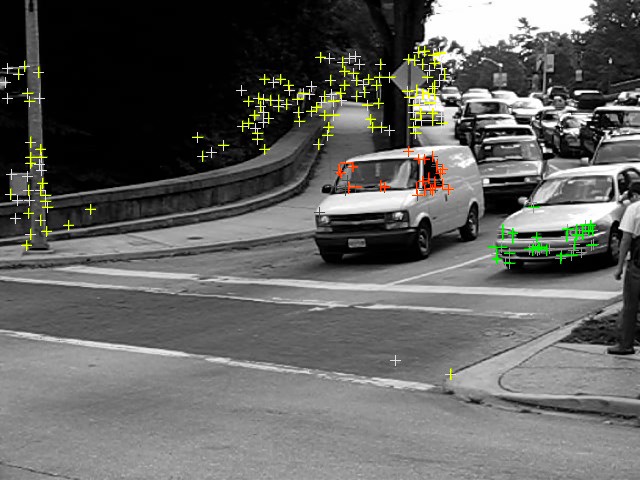}    \label{fig:hopkin_example}}
\vspace*{-0.1cm}                           
\caption{Real data. (a): Faces from the Extended Yale Dataset B. (b): Images from the Hopkins 155 dataset.}
\vspace*{-0.2cm}
\end{figure}
\vspace*{-0.2cm}

\subsection{ Real Data: Motion Segmentation}
Motion segmentation of trajectory data has been widely studied in computer vision applications such as reconstructing dynamic scenes \cite{yan2006general}.
In dynamic scenes with multiple moving rigid objects, the trajectories can be represented by high-dimensional vectors. Yet, they can span low-dimensional linear manifolds \cite{yan2006general}. It is shown that, in an affine subspace, trajectories of a single motion in an ambient dimension $ \mathbb{R}^{2F}$, where $F$ is the number of frames, lies in a low-dimensional linear subspace of up to four dimensions.  
Thus, subspace clustering algorithms can be used to cluster the trajectories of different motions in separate subspaces.
In this part, we used Hopkins 155 motion segmentation dataset \cite{tron2007benchmark} to examine the proposed subspace clustering method. %Figure \ref{fig:hopkin_example} shows examples from this dataset.
 We compared the misclassification errors of our proposed method on this dataset with LRR\cite{liu2010robust}, SSC \cite{elhamifar2009sparse}, and S$^3$C \cite{li2017structured}. 
Table \ref{table:HopkinsErr} shows the clustering error percentages for these methods. We cite the reported results in   \cite{li2017structured} in this table. As shown, our proposed method outperforms all other state-of-the-art algorithms.

%\vspace*{-0.2cm}                           
\begin{table}[ht]
\centering                                            
\caption{Average $\%$misclassification on Hopkins dataset.}   
\vspace*{-0.2cm}        
\begin{tabular}{|c|c|c|c|c|c|c|}   
\hline 
 & \multicolumn{2}{c|}{$2$ Motions} & \multicolumn{2}{c|}{$3$ Motions} & \multicolumn{2}{c|}{Total}\\
$\%$Err & Avg & Median &  Avg  & Median & Avg &Median \\
 \hline
 \hline
LRR & 3.76 & 	\textbf{0.00} & 9.92 & 1.42 		& 5.15 &\textbf{0.00}\\
LRSC & 2.57      & 0.00      & 6.62  & 1.76   & 3.47 & 0.09   \\
SSC & 1.95 & 	\textbf{0.00} & 4.94 & {0.89}& 2.63 & \textbf{0.00}\\
S$^3$C & 1.73 & \textbf{0.00} & 5.50 & {0.81}& 2.58 &\textbf{0.00 }\\
Soft-S$^3$C & 1.65 & \textbf{0.00} & 4.27 & {0.61}& 2.24 &\textbf{0.00 }\\
\hline
\hline
Prob SSC & \textbf{1.57}  & \textbf{0.00 }  &  \textbf{3.60}  & \textbf{0.58} & \textbf{2.13} &\textbf{0.00 }\\
\hline  
\end{tabular}                    
%\end{adjustbox}
\label{table:HopkinsErr}  
\end{table}

\section{Conclusion} \label{sec:conclude}

We introduced a new subspace clustering method that outperforms state-of-the-art methods reported recently in the literature. This boost in accuracy is because we replace the usual clustering matrix $Q$ with an {\em association matrix} $\mathit{A}$ that allows us to track the assignment of points in the same clusters, and hence delay hard assignments until later iterations, when more confidence is gained. This is possible because, at each iteration, the method splits the points into two groups of {\em certain} and {\em uncertain}, allowing the latter group's association to be delayed until later iterations when the association probabilities become higher. A direct advantage of this delayed association is that the method performs better when subspaces are highly overlapping (i.e. high intersection of bases). The results on both synthetic and real data confirm these advantages. %Using incremental spectral clustering can reduce the time complexity because we can apply it to {\em uncertain} points and their connections to the rest of the data. However, when we have a high rate of {\em uncertain} points, we need to re-initialize the clusters using classical spectral clustering methods to prevent growth of inaccuracy.

%-------------------------------------------------------------------------

%
%
%{\small
%\bibliographystyle{aaai}
%\clearpage

{\small
\bibliographystyle{IEEEtran}
\bibliography{egbib}
}

\end{document}